# Pattern Generation Strategies for Improving Recognition of Handwritten Mathematical Expressions

Anh Duc LE , Bipin Indurkhya, and Masaki NAKAGAWA

## Abstract

Recognition of Handwritten Mathematical Expressions (HMEs) is a challenging problem because of the ambiguity and complexity of two-dimensional handwriting. Moreover, the lack of large training data is a serious issue, especially for academic recognition systems. In this paper, we propose pattern generation strategies that generate shape and structural variations to improve the performance of recognition systems based on a small training set. For data generation, we employ the public databases: CROHME 2014 and 2016 of online HMEs. The first strategy employs local and global distortions to generate shape variations. The second strategy decomposes an online HME into sub-online HMEs to get more structural variations. The hybrid strategy combines both these strategies to maximize shape and structural variations. The generated online HMEs are converted to images for offline HME recognition. We tested our strategies in an end-to-end recognition system constructed from a recent deep learning model: Convolutional Neural Network and attention-based encoder-decoder. The results of experiments on the CROHME 2014 and 2016 databases demonstrate the superiority and effectiveness of our strategies: our hybrid strategy achieved classification rates of 48.78% and 45.60%, respectively, on these databases. These results are competitive compared to others reported in recent literature. Our generated datasets are openly available for research community and constitute a useful resource for the HME recognition research in future.

## 1 Introduction

Mathematical Expressions (MEs) are used often in education, science, engineering, business and even in daily life. When creating a document with MEs, there are basically three methods to input MEs to a computer: 1) using a math description language like LATEX, 2) employing an equation editor like Microsoft Equation Editor, or 3) using handwritten ME recognition. The first two methods are popular among scientists, engineers, and other professionals. For ordinary users, it is difficult to remember the grammar of math symbols/expressions in LATEX or to search a menu containing a long list of expression templates. Handwritten ME recognition can overcome these problems. Users can just write an ME on a touch screen and the handwritten ME recognizer will translate it to LATEX or MEE automatically.

Recognition of Handwritten Mathematical Expressions (HMEs) is one of the current challenges facing handwriting recognition because of the ambiguity and complexity of two-dimensional (2D) patterns. These challenges include both online HME recognition (input handwritten pattern is a time sequence of sampled pen points) and offline HME recognition (input handwritten pattern is an image). Traditional recognition methods for both online and offline HMEs proceed in three stages. First, a sequence of input strokes or an input image is segmented into potential symbols (symbol segmentation). Then, these potential symbols are recognized by a symbol classifier (symbol recognition). Finally, structural relations among the recognized symbols are determined and the structure of the expression is analyzed by a parsing algorithm in order to provide the most likely interpretation of the input HME (structural analysis). The recognition problem requires not only segmenting and recognizing symbols but also analyzing 2D structures and interpreting structural relations. Recognition of offline HMEs is more challenging than that of online HMEs because it lacks time sequence information, which makes symbol segmentation more difficult and incurs more computation time for structural analysis. Inspired by the recent successes of the attention-based encoder-decoder model in neural machine translation [1] and image caption generation [2], some researchers have applied this model for offline HME recognition. It requires only HME images and their corresponding codes in Latex for training. Symbol segmentation/recognition and structural analysis are incorporated in attention-based encoder-decoder models. This method outperformed traditional methods. However, it required a large amount of data and a powerful machine (GPU) for training.

Many approaches have been proposed for recognizing online HMEs during the last two decades, as summarized in survey papers [3, 4] and recent competition papers [5-7]. Most of these employed the traditional method mentioned above for online HME recognition, whereas some recent works [8 - 11] applied the attention-based encoder-decoder that was developed for offline HMEs. In the following, we review a few recent approaches developed in academia and industry that are evaluated on the recent Competition on Recognition of Online Handwritten Mathematical Expressions (CROHME).

## 1.1 Academic systems:

A system for recognizing online HMEs using a top-down parsing algorithm was proposed by MacLean et al. [12]. The incremental parsing process constructs a shared parse forest that presents all recognizable parses of the input. Then, the extraction process finds the top to $n^{th}$-most highly ranked trees from the forest. By using the horizontal and the vertical orderings, this method reduces infeasible partitions and dependence on the stroke orders. However, in the worst-case, the number of sub-partitions that must be considered during parsing and the complexity of the parsing algorithm is still quite large, as $O(n^4)$ and $O(n^4|P|)$, respectively. This system participated in CROHME 2012.

A global approach allowing mathematical symbols and structural relations to be learned directly from expressions was proposed by Awal et al. [13]. During the training phase, symbol hypotheses are generated without using a language model. The dynamic programming algorithm finds the best segmentation and recognition of the input. The classifier learns both the correct and incorrect segmentations. The training process is repeated to update the classifier until it recognizes the training set of online HMEs correctly. Furthermore, contextual modeling based on structural analysis of online HMEs is employed, where symbol segmentation and classification are learnt directly from expressions using the global learning scheme. This system participated in all competitions.

A formal model for online HME recognition based on 2D Stochastic Context Free Grammar (SCFG) and Hidden Markov Model (HMM) was proposed by Alvaro et al. [14]. HMM uses both online and offline features to recognize mathematical symbols. The Cocke-Younger-Kasami (CYK) algorithm is modified to parse an input online HME in two dimensions (2D). They use the range search to improve time complexity from $O(n^4|P|)$ to $O(n^3 logn|P|)$. To determine structural relations among symbols and subexpressions, a Support Vector Machine (SVM) learns geometric features between bounding boxes. This system was judged to be the best system at CROHME 2011 and the best system with using only CROHME data at CROHME 2013 and 2014.

Le et al. presented a recognition method for online HMEs based on SCFG [15]. Stroke order is exploited to reduce the search space and the CYK algorithm is used to parse a sequence of input strokes. Therefore, the complexity of the parsing algorithm is reduced, but is still $O(n^3|P|)$, which is similar to the original CYK algorithm. They extended the grammar rules to cope with multiple symbol order variations and proposed a concept of body box with two SVM models for classifying structural relations. The experiments showed a good recognition rate and reasonable processing time. The system participated in CROHME 2013, 2014, and 2016.

A parsing algorithm based on a modified version of the Minimum Spanning Tree (MST) was presented by Hu et al. [16]. The parser extracts MST from a directed Line-of-Sight graph. The time complexity of this algorithm is lower than the time complexity of the CYK method. This parser achieved a good result for structure analysis on online HME patterns assuming a correct segmentation and symbol recognition. This system participated in all competitions.

Zhang et al. proposed a method using a Bidirectional Long Short-Term Memory neural network (BLSTM) for interpreting 2D languages such as online HMEs [17]. This is an end-to-end model, which requires as input only online HMEs and their corresponding codes in LaTex or MathML, and is able to produce results using a BLSTM model. However, the performance is low due to stroke order variations.

Recently, an attention-based encoder-decoder approach has been successful in machine translation [1] and image caption generation [2]. It outperforms traditional methods in many sequence-to-sequence tasks. Deng et al. extended this approach to recognize offline images of printed MEs and HMEs [8], and called it WYSIWYG (What You See Is What You Get). This recognition system gives an encouraging result on offline recognition of printed MEs and HMEs.

Zhang presented an end-to-end approach based on neural network to recognize offline HMEs, which is called WAP [9]. They employed a convolutional neural network encoder to extract features from an input offline HME and a recurrent neural network decoder with an attention-based parser to generate LaTeX sequences. For improvement accuracy, they use a deep Convolutional Neural Network (CNN) containing 16 convolutional layers and a coverage model to avoid over- or under-parsing problems. This system achieved the best expression recognition accuracy on CROHME 2014 test set using the official training data at the time of its publication. Later, an improvement of WAP was achieved by employing Multi-Scale Attention with DenseNet encoder [10].

In our previous work, we proposed local and global distortion models for generating online HMEs from the CROHME database. In this approach, online HMEs are converted to images for training an end-to-end recognition system [11], which has three parts: a convolution neural network for feature extraction, a BLSTM for encoding extracted features, and an LSTM and an attention mechanism for generating target LaTex. Our experimental results confirmed the efficacy of generated data.

### 1.2 Industrial Systems:

The MyScript math recognition system handles segmentation, recognition and interpretation concurrently in order to produce the best candidates. This system uses a 2D grammar to analyze the spatial relationships between all parts of the equation to determine the segmentation of all of its parts. The symbol recognizer extracts both online and offline features from the pen trajectory and its converted image, respectively. Online features include direction and curvature. Offline features are typically based on projections and histograms. The features are processed by a combination of Deep MLP and Recurrent Neural Networks. The recognizer also includes a statistical language model that evaluates contextual probabilities between the symbols in the equation. It is trained on about 30,000 additional handwritten samples collected from writers in different countries. It received the best system prize from CROHME 2012 to 2016 [5 - 7].

The WIRIS is a math recognition system based on an integrated approach such that symbol segmentation, symbol classification and the structure of the math expression are globally determined. It is an improved version of Alvaro's recognition system. A probabilistic grammar guides the recognition process to calculate the structural probabilities between symbols and expressions. The system is integrated with a statistical language model estimated from the expressions provided by the competition organizers. The system achieved the best performance using only the CROHME data at CROHME 2016 [7].

## 1.3 Database

Mouchere et al. have been organizing CROHME for fair evaluation based on common databases. The organizer increased the training set from CROHME 2011 to 2012. Then, they combined the CROHME 2012 training set with five other datasets of online HMEs and made it available as a training set for later competitions. [5]. The CROHME training set is useful, but it currently contains only 8835 online HMEs. It is very small in comparison with the training data for machine translation (WMT14) [18] and those for image captioning (Microsoft COCO) [19]. Increasing the number of online HMEs is very hard because it takes time and effort to collect handwriting and generate ground-truths for them at different levels such as stroke, symbol, and structure. Data synthesis is one solution for generating more data from a small training set like CROHME. Leung et al. and Cheng et al. proposed distortion models to generate more training data, respectively, for Chinese and Japanese character recognition [20, 21]. Plamondon et al. proposed delta-lognormal model for handwriting analysis [22], which can be used for data augmentation [23]. The above data synthesis methods make variations only in the shape of the handwriting. For HME, however, we need a synthesis method which makes variations with respect to both the shape and the structure.

In this paper, we propose data generation strategies for HMEs. We employ online HMEs since many public databases are available but the strategies are also effective for offline HMEs with slight modifications. The first strategy employs distortion models to generate variations of shape for both symbols and HMEs as in our previous work [11]. The second strategy employs an HME decomposition model to generate more structural ME variations. The third strategy is a hybrid model, which generates HMEs by the decomposition model and then distorts them by the distortion models. The hybrid strategy enriches shapes and structures for the training data. We apply these strategies for online HMEs, but then convert online HMEs to offline HMEs (images). Then, we train our end-to-end HME recognition system, which is a modified version of our previous end-to-end offline recognition system [11] with feature extraction (deeper CNN than the previous system and dropout technique) added. To compare our system with Deng’s recognition system [8], we employ a deep CNN with dropout for better generalization of feature extraction. To compare our system with Zhang’s recognition system [9, 10], we employ a simpler decoder, which contains the LSTM decoder and the basic attention mechanism.

We have shown that the recognition accuracy can be improved significantly just by enriching the shapes and structures of the existing dataset, in contrast with the more sophisticated approaches, such as the one provided by Zhang [9, 10], with an advanced architecture and training process for the recognition system by integrating a dense network, coverage technique for attention, multi-scale attention, weight noise for regularization and an ensemble method.

The rest of this paper is organized as follows. The distortion, decomposition, and hybrid strategies for generating more HMEs are described in Section 2. An end-to-end recognition system for offline HMEs is presented in Section 3. The experimental results are presented and discussed in Section 4, and the final conclusions are presented in Section 5.

# 2 PATTERN GENERATION STRATEGIES

Inspired by recent successes of deep learning in handwriting recognition and machine translation, supported by a huge number of labeled data, we propose pattern generation strategies to improve the performance of the end-to-end HME recognition from training data provided by the CROHME competition.

## 2.1 HME Distortion Strategy

In this approach, we extend Cheng et al.’s model [21] into local and global distortions of HMEs. Local distortions are applied for symbols, while global distortions are applied for the whole HMEs. Local distortions include shear, shrink, perspective, shrink plus rotation, and perspective plus rotation. Global distortions include scaling and rotation. The distortion process is shown in Figure 2. First, all the symbols in an HME are distorted by the same distortion model. Then, the HME is further distorted by scaling and rotation models sequentially. The distortion models are described below.

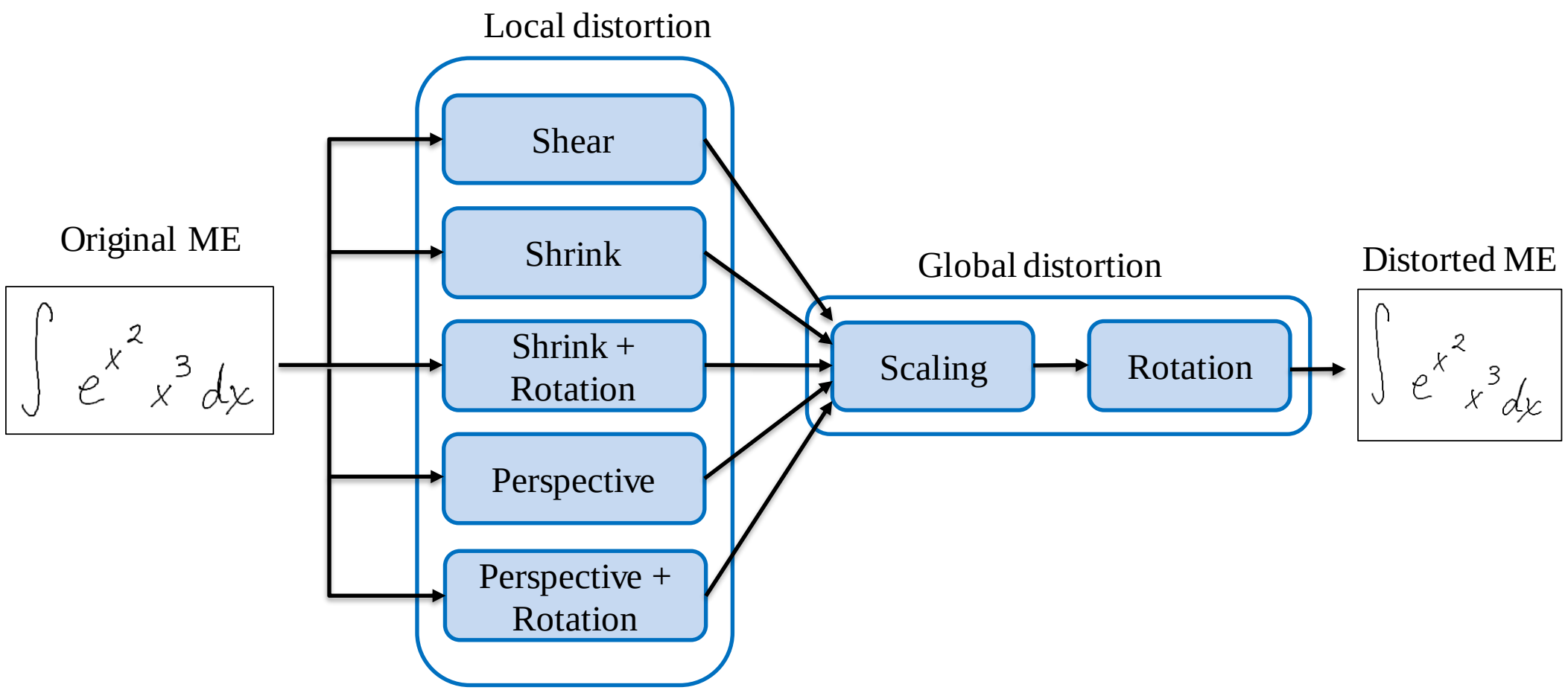

**Fig. 1.** Process of distortion model for patterns generation.

### 2.1.1 *Local Distortion*

Shear is a transformation that moves points on one axis by a distance that increases linearly in proportion to the distance with the other axis. Shear can be along horizontal or vertical axes, and is calculated by Eqs. (1) and (2).

$$\begin{cases} x' = x + ytan\alpha \\ y' = y \end{cases} \quad (1) \qquad \begin{cases} x' = x \\ y' = y + xtan\alpha \end{cases} \quad (2)$$

Shrink and Perspective are both similar to Shear with different transformation equations. The vertical and horizontal shrink models are described in Eqs. (3) and (4), respectively; and the vertical and horizontal perspective models are shown in Eqs. (5) and (6), respectively.

$$\begin{cases} x' = y\left(sin\left(\frac{\pi}{2} - \alpha\right) - \left(\frac{xsin(\alpha)}{100}\right)\right) \\ y' = y \end{cases} \quad (3) \qquad \begin{cases} x' = x \\ y' = x\left(sin\left(\frac{\pi}{2} - \alpha\right) - \left(\frac{ysin(\alpha)}{100}\right)\right) \end{cases} \quad (4)$$

$$\begin{cases} x' = \frac{2}{3}\left(x + 50cos\left(4\alpha\frac{x-50}{100}\right)\right) \\ y' = \frac{2}{3}y\left(sin\left(\frac{\pi}{2} - \alpha\right) - \left(\frac{ysin(\alpha)}{100}\right)\right) \end{cases} \quad (5) \qquad \begin{cases} x' = \frac{2}{3}x\left(sin\left(\frac{\pi}{2} - \alpha\right) - \left(\frac{xsin(\alpha)}{100}\right)\right) \\ y' = \frac{2}{3}\left(y + 50cos\left(4\alpha\frac{y-50}{100}\right)\right) \end{cases} \quad (6)$$

The shrink plus rotation model applies these two models sequentially, and the perspective plus rotation model is similar. The rotation model is shown in Eq. (7).

$$\begin{cases} x' = xcos\beta + ysin\beta \\ y' = xsin\beta + ycos\beta \end{cases} \quad (7)$$

where $(x, y)$ is the coordinate of a pen point and $(x', y')$ is the new coordinate transformed by a local distortion model, $\alpha$ is the angle of the shear, shrink, and perspective distortion models and $\beta$ is the angle of the rotation distortion model. The local distortion models and their parameters are presented by $(id, \alpha, \beta)$, where *id* is the identifier of the distortion model from 1 to 5, and $\alpha$ and $\beta$ are from $-10^{o}$ to $10^{o}$. The above definitions of distortion are for online patterns, but similar definitions can be made for offline patterns by interpreting $(x, y)$ as the coordinates of a black pixel.

Figure 3 show examples of the local distortion models with $\alpha = 10^{o}$ and $\beta = 10^{o}$.

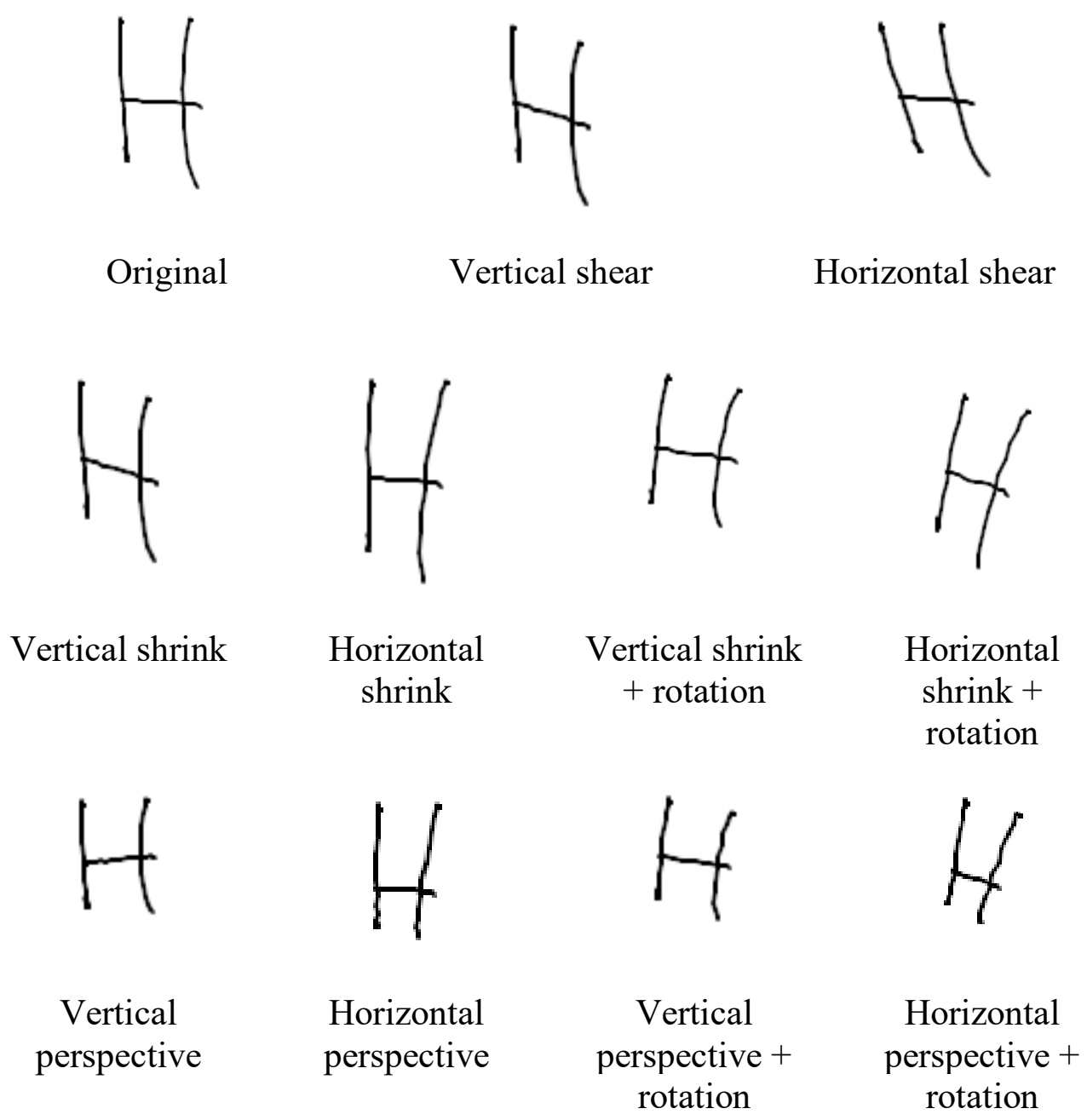

**Fig. 2.** Examples of local distortion by shear, shrink and perspective transformations.

### *2.1.2 Global Distortion*

Global distortion transforms an HME by using rotation and scaling sequentially. The rotation model is calculated by Eq. (8) and the scaling model is shown in Eq. (9).

$$\begin{cases} x' = x\cos\gamma + y\sin\gamma \\ y' = x\sin\gamma + y\cos\gamma \end{cases} \quad (8) \qquad \begin{cases} x' = kx \\ y' = ky \end{cases} \quad (9)$$

where $k$ is the scaling factor. The parameters of the global distortion model are presented by $(k, \gamma)$ where $\gamma$ is the angle of the global rotation distortion model, $k$ is from 0.7 to 1.3, and $\gamma$ is from $-10^{o}$ to $10^{o}$. An example of the global distortion is shown in Figure 4. Global distortion of offline patterns can be defined similarly as for local distortion described above.

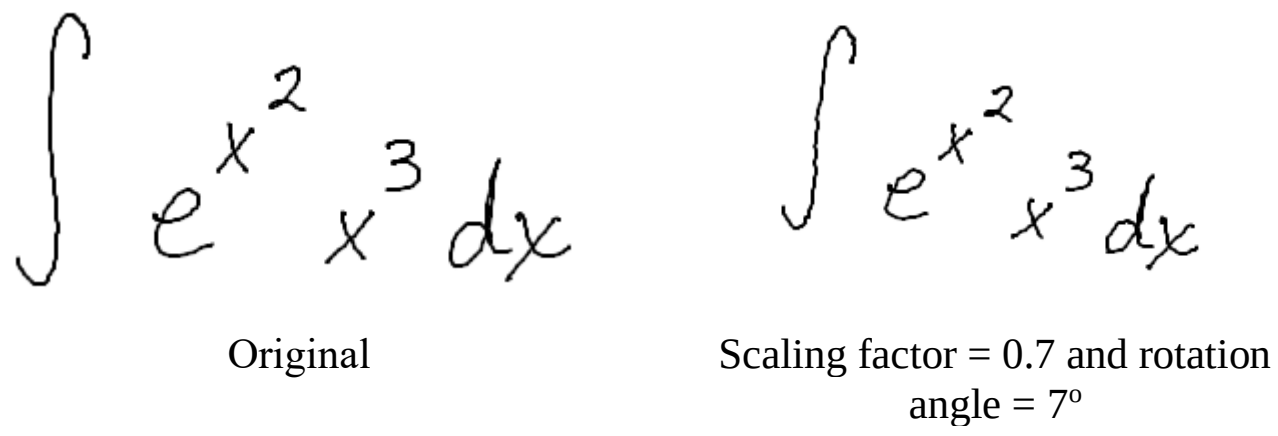

**Fig. 3.** Examples of global distortion by scaling and rotation models.

### *2.1.3 Pattern generation*

To generate an HME, we first randomize five variables $(id, \alpha, \beta, k, \gamma)$. Then, all the symbols in an HME are distorted by local distortion models with $(id, \alpha, \beta)$. Then, the HME is further distorted by the global distortion model with $(k, \gamma)$. Figure 5 shows some generated HMEs from the original HME that appeared in Figure 3.

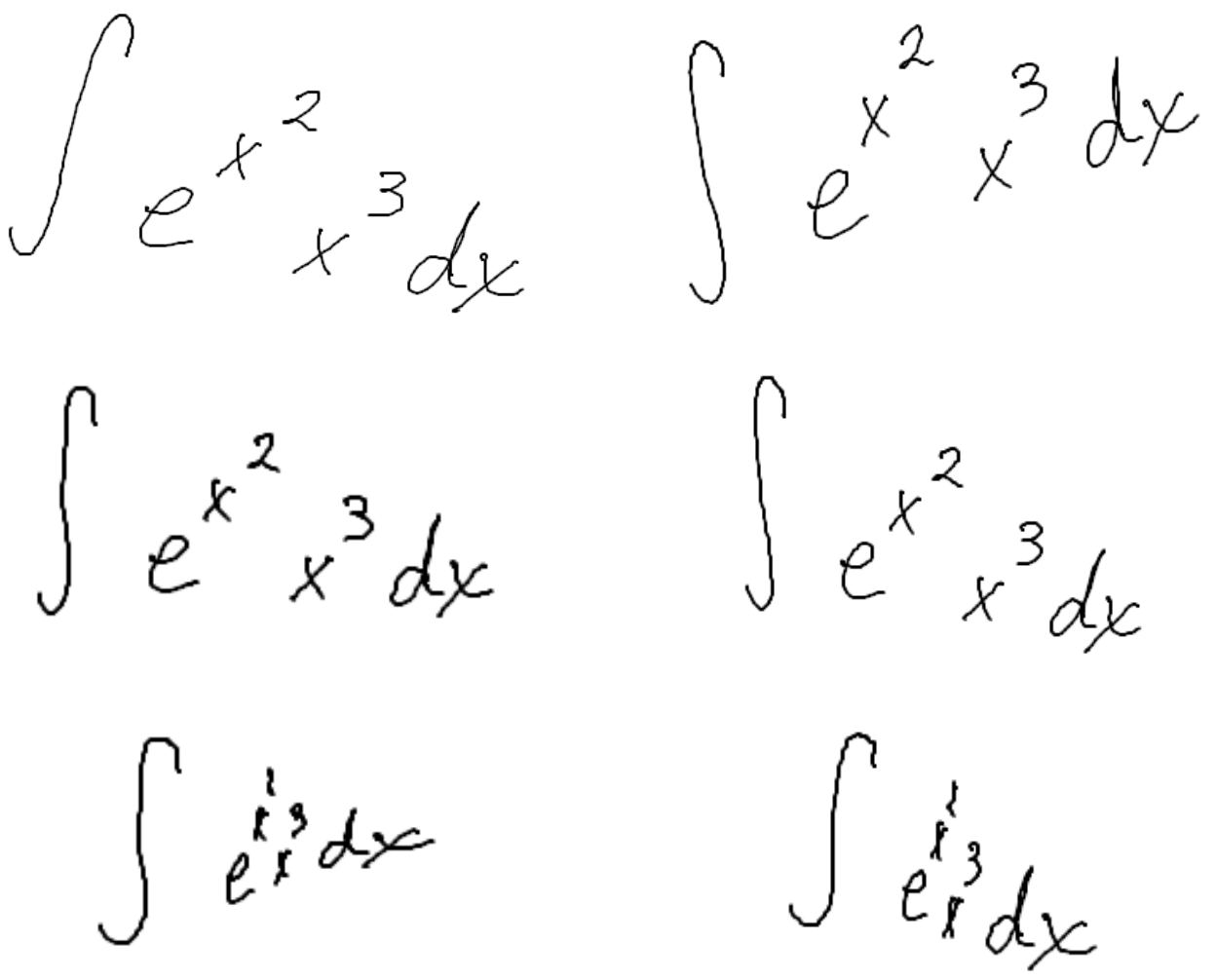

**Fig. 4.** Sample OHMEs generated by combination of local and global distortion models.

## 2.2 HME decomposition strategy

An HME is constructed by combining symbols/sub-HMEs under ME grammar rules, which constitute a context-free grammar. For example, an HME for $x^2 + 1$ is constructed from that for $x^2$ and a handwritten symbol *1* under the grammar rule: ME → ME + SYM. Therefore, we can decompose an HME to sub-HMEs to get more HME patterns. Given an existing dataset, we can expand it by decomposing all HMEs in the dataset to sub-HMEs. For decomposing an HME, we construct a Symbol Relation Tree (SRT) and extract sub-HMEs from the SRT by the following rules:

- Rule 1: If an HME contains subscript/superscript, the new sub-HME is generated by removing all the subscript/superscript parts. This is the baseline sub-HME.
- Rule 2: Sub-HMEs are created from the subscript/superscript/over/under/inside parts in the SRT.
- Rule 3: For every binary operator in the baseline, we generate two sub-HMEs from the left and right parts of the operator. This rule is not applied to binary operators inside brackets to avoid generating invalid HMEs.
- Rule 4: To limit the number of HMEs, we discard sub-HMEs having only one symbol.

Figure 5 presents an example of the decomposition strategy. From the HME $x^2 + 2x + 1$, we generate the sub-HMEs: $x + 2x + 1$ (Rule 1), *2* (Rule 2), $x^2$, $2x + 1$, $x^2 + 2x$, *1* (Rule 3). Then, we apply Rule 4 to discard sub-HMEs *2* and *1*.

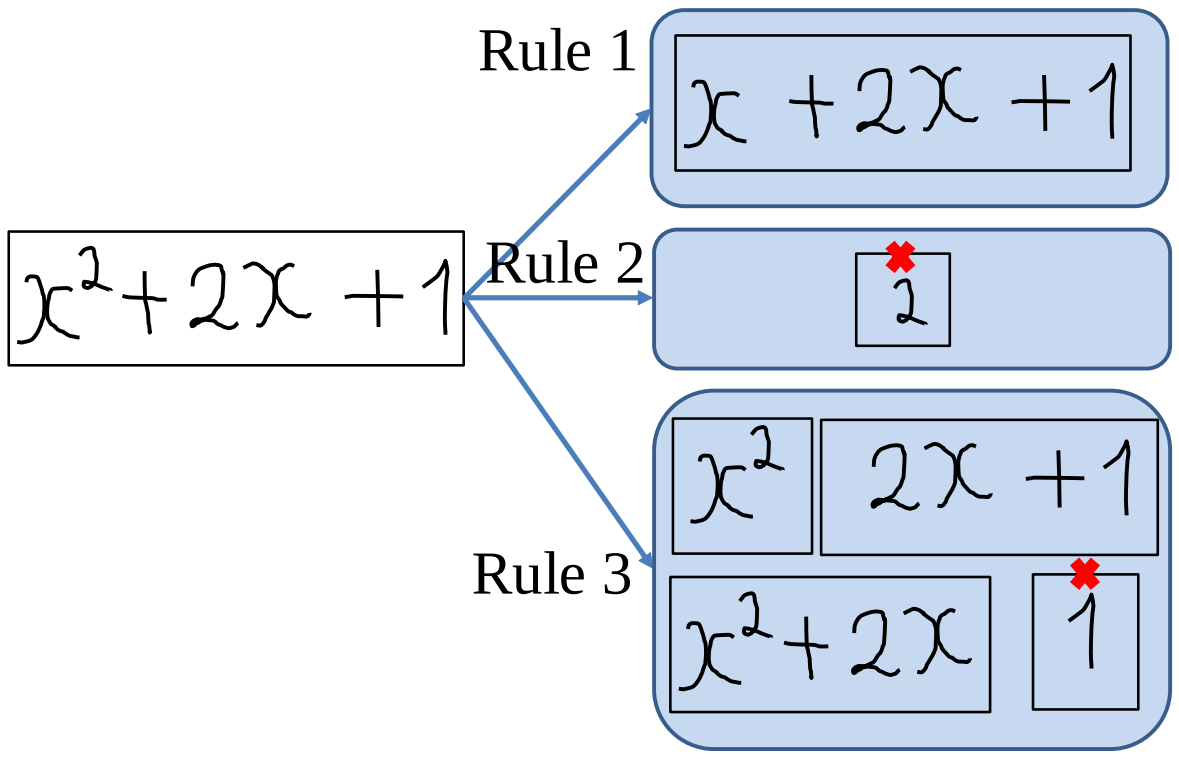

**Fig. 5.** An example of HME decomposition strategy. (sub-HMEs with red X are discarded by Rule 4 )

## 2.3 Hybrid strategy

For the hybrid strategy, we first employ the HME decomposition strategy to generate sub-HMEs. Then, we employ the local and global distortion strategies for each sub-HME to generate distorted sub-HMEs. By employing all the three strategies, we introduce wider variations in the training data with respect to both shape and structure.

# 3 OVERVIEW OF THE END-TO-END RECOGNITION SYSTEM

The architecture of our end-to-end recognition system is shown in Figure 6. It has three modules: a convolution neural network for feature extraction from an offline HME, a BLSTM Row Encoder for encoding extracted features, and an LSTM Decoder and an attention model for generating the target LaTex. These modules are described in turn in the following sections.

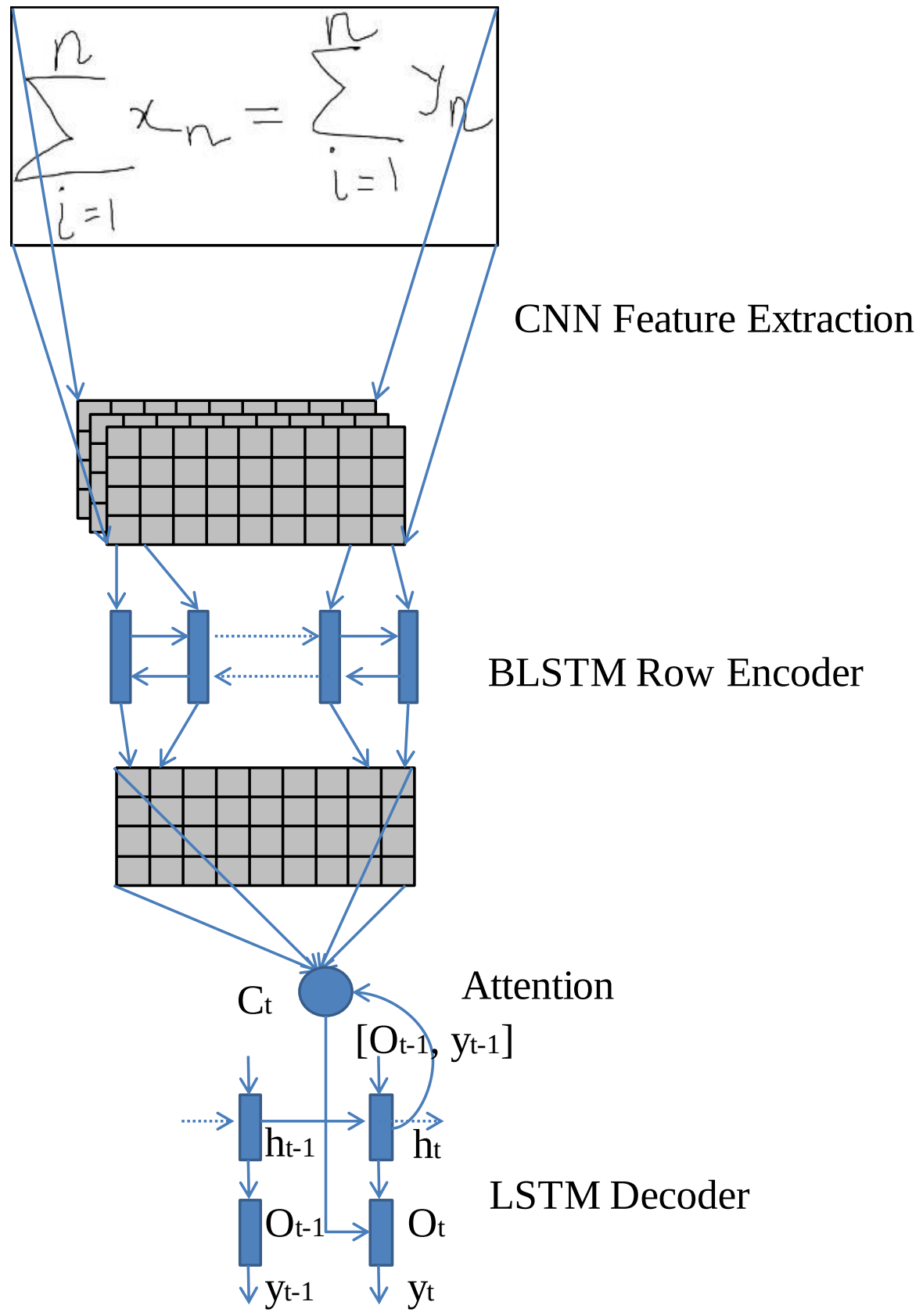

**Fig. 6.** Architecture of the end-to-end recognition system

## 3.1 Feature Extraction by CNN

Features are extracted from an offline HME by a convolution neural network containing multiple layers of convolution and max-pooling layers. We use recent techniques such as dropout and batch normalization to improve feature extraction. The details of the CNN module are described in the experimental section. Using the CNN feature extraction, an input image ($H$ x $W$) is transformed to a feature grid $F$ of size ($K$ x $L$ x $D$), where ($H$, $W$), ($K$, $L$), and $D$, respectively, are the image size, the reduced size, and the feature size.

## 3.2 Row Encoder

For the row encoder, we employed a model similar to the one proposed by Deng et al. [8]. A Bidirectional Long Short-Term Memory neural network (BLSTM) is applied to each row to extract new row features. The BLSTM encodes a feature grid *F* into a new feature grid *F'*. For $u \in [1, \ldots K]$ and $v \in [1, \ldots L]$, the new feature $F'_{u,v}$ is calculated by the following equations:

$$\overrightarrow{F'_{u,v}} = \overrightarrow{LSTM}(\overrightarrow{F'_{u,v-1}}, F_{u,v}) \quad (8)$$

$$\overleftarrow{F'_{u,v}} = \overleftarrow{LSTM}(\overleftarrow{F'_{u,v+1}}, F_{u,v}) \quad (9)$$

$$F'_{u,v} = \overrightarrow{W} * \overrightarrow{F'_{u,v}} + \overleftarrow{W} * \overleftarrow{F'_{u,v}} + B \quad (10)$$

where $\overrightarrow{LSTM}$, $\overleftarrow{LSTM}$, $\overrightarrow{F'_{u,v}}$, and $\overleftarrow{F'_{u,v}}$, respectively, are the forward and backward LSTM and their outputs at $(u, v)$; and $\overrightarrow{W}$, $\overleftarrow{W}$, and *B* are the weight matrices of the forward, backward LSTMs and the bias vector.

## 3.3 Decoder

The decoder generates one symbol at a time. At each time step *t*, the probability of decoding each word $y_t$ is calculated as:

$$p(y_t | y_1, \ldots, y_{t-1}, F') = softmax(W_{out} * O_t) \quad (11)$$

$$O_t = \tanh(W_c * [h_t ; c_t]) \quad (12)$$

$$h_t = LSTM(h_{t-1}, [y_{t-1}, O_{t-1}]) \quad (13)$$

$$c_t = \sum_{u,v} \alpha_{u,v} * F'_{u,v} \quad (14)$$

$$\alpha_{u,v} = \mathrm{softmax}(\tanh(\mathrm{W}_h * \mathrm{h}_t + \mathrm{W}_{F'} * \mathrm{F'}_{u,v})) \quad (15)$$

where $h_t$ and $O_t$, respectively, are the hidden unit and the output of the LSTM decoder at time step *t*; and $c_t$ is a context vector computed as the weighted sum of the feature grid *F'* and their weights $\alpha$ . We follow the attention model proposed by Luong et al.[24] to calculate $\alpha_{u,v}$ as per Equation 15.

## 3.4 Training:

We employed cross-entropy as an objective function to maximize the probability of predicted symbols as follows:

$$f = \sum_{i=1}^{|D|} \sum_{t=1}^{|HME_i|} -\log p(g_{i,t} | y_1, \ldots, y_{t-1}, F'_i) \quad (16)$$

where *D* is the training ME corpus, $HME_i$ is a training sample and $g_{i,t}$ is a ground truth latex symbol.

We used mini-batch stochastic gradient descent to learn the parameters. The initial learning rate was set to 0.1. The training process was stopped when the recognition rate on the validation set did not improve after 10 epochs. We tried several settings of the CNN feature extraction but with fixed settings of the encoder and decoder. The hidden states of the encoder and decoder are 256 and 512, respectively. The system was implemented on Torch and the Seq2seq-attn NMT system [25]. All the experiments are performed on a 4GB Nvidia Tesla K20.

# 4 EVALUATION

We employed the CROHME training dataset and three generated datasets for training, the CROHME 2013 test set for validation, and the CROHME 2014 as well as CROHME 2016 test sets for testing. The CROHME training dataset contains 8,835 online HMEs for training, and the CROHME 2013, 2014 and 2016 testing datasets contain 671 online HMEs, 986 online HMEs and 1127 online HMEs, respectively. The number of symbol classes is 101.

First, we trained the end-to-end system by varying the settings of the CNN feature extraction to determine the best setting, which is used in the end-to-end system. For data generation, we created three new training datasets from the CROHME training dataset using three pattern generation strategies: distortion, decomposition, and hybrid. To avoid overfitting, we employed global distortion as data augmentation during the training phase. Finally, we compared the performances of the end-to-end system trained on different datasets with other systems which participated in CROHME 2014 and 2016.

## 4.1 Training datasets

We generated additional HME patterns using the three strategies mentioned above. In the distortion strategy, we generated five new online HMEs from each original online HME. We call it this dataset CROHME-Distortion. In the decomposition strategy, we generated new sub online HMEs from the original training set; this is referred to as CROHME-Decomposition. In the hybrid strategy, we applied the distortion technique to generate five new online HMEs for each online HME on CROHME-Decomposition. This dataset is referred to as CROHME-Hybrid. The generated datasets also include original online HMEs from the CROHME training set. The number of online HMEs and generated online HMEs for each training set are shown in Table I. All online HMEs were converted to offline HMEs for training the recognition system by connecting pen points with constant thickness of 3 pixels wide. All the datasets are openly available for research [26].

TABLE I. DESCRIPTION OF TRAINING SETS

| **Training set** | **CROHME training set** | **CROHME-Distortion** | **CROHME-Decomposition** | **CROHME-Hybrid** |
|---|---|---|---|---|
| **# of HMEs** | 8,835 | 53,010 (8,835x5+8,835) | 32,884 (24,049+8,835) | 197,304 (32,884x5+32,884) |

## 4.2 Determination of the structure of CNN feature extraction

The first experiment was for selecting the best CNN feature extraction. Since the structure of CNN feature extraction affects the performance of the end-to-end system [8 - 11], we tried several settings of CNN to train the end-to-end system on the CROHME training dataset. Three variations of VGG model [27], as shown in Table II, were created by changing the number of layers, filters in each layer, and dropout rate for each layer. The best combination was selected based on the accuracy on CROHME 2013, which was CNN3 with 13 Conv.-Norm-ReLU layers and 4 max-pooling layers. Dropout with the drop rate 0.2 was applied on the last 7 Conv.-Norm-ReLU layers.

## 4.3 Evaluation of data generation strategies

The second experiment evaluated the performance of the end-to-end recognition systems trained on the CROHME training set, CROHME-Distortion, CROHME-Decomposition, and CROHME-Hybrid. We used the CROHME 2013 testing dataset for validation. An offline HME was judged to be recognized correctly in terms of expression level if all of its symbols, relations and structure were recognized correctly. For measurement, we used the expression recognition rate (ExpRate), which counts offline HMEs recognized at the expression level over all the testing offline HMEs. Table III shows the recognition rates of the end-to-end systems trained on different training sets on the CROHME 2014 and CROHME 2016 test set. We can see that ExpRate increases when employing our proposed data generation strategies. The recognition system trained on CROHME-hybrid archived the best performance on CROHME 2014 (48.78%) and CROHME 2016 (45.60%). We employed paired t-test to compare the performance of data generation strategies. The recognition of $n$ offline HMEs can be assumed as Bernoulli trials and the probability of misrecognition is $p$. Therefore, the mean and variance were calculated as $np$ and $np(1 - np)$, respectively. The significant differences between the pair recognition systems trained on (CROHME-Decomposition vs CROHME), (CROHME-hybrid vs CROHME) were verified, while the significance was not verified for (CROHME-Distortion vs CROHME), (CROHME-hybrid, CROHME-Decomposition) on the CROHME test set 2014 and 2016 with $P < 0.01$. The recognition accuracy was improved significantly on datasets generated by decomposition strategies. The recognition accuracy on CROHME-Decomposition was better than that on CROHME-Distortion (47.67% vs 41.48% on CROHME 2014 test set and 42.63% vs 37.14% on CROHME 2016 test set), although the number of patterns of CROHME-Decomposition was smaller than that of CROHME-Distortion. Since the current training with CROHME still lacks structural variations, it is not enough for end-to-end recognition systems to learn. The decomposition strategy enlarges structural variations, resulting in a significant improvement in the performance. Thus, we conclude that our proposed decomposition strategy is useful and effective.

TABLE II. STRUCTURES OF CNN FEATURE EXTRACTION. THE PARAMETERS OF THE CONVOLUTION AND MAX POOLING LAYERS ARE DENOTED AS "CONV.-NORM-RELU: FILTER SIZE (NUMBER OF FILTERS)" AND "MAX POOL FILTER SIZE", RESPECTIVELY.

| | **CNN1** | **CNN2** | **CNN3** |
|---|---|---|---|
| **Structure** | Conv-ReLU:3x3 (64)<br>Max-pool 2x2<br>Conv-ReLU:3x3 (128)<br>Max-pool 2x2<br>Conv-Norm-ReLU:3x3 (256)<br>Conv-ReLU:3x3 (256 )<br>Max-pool 1x2<br>Conv-Norm-ReLU:3x3 (512 )<br>Max-pool 2x1<br>Conv-Norm-ReLU:3x3 (512 ) | Conv-Norm-ReLU:3x3 (50)<br>Conv-Norm-ReLU:3x3 (100)<br>Max-pool 2x2<br>Conv-Norm-ReLU:3x3 (150)<br>Conv-Norm-ReLU:3x3 (200)<br>Max-pool 2x2<br>Conv-Norm-ReLU:3x3 (250) drop 0.2<br>Conv-Norm-ReLU:3x3 (300) drop 0.2<br>Max-pool 1x2<br>Conv-Norm-ReLU:3x3 (350) drop 0.2<br>Conv-Norm-ReLU:3x3 (400) drop 0.2<br>Max-pool 2x1<br>Conv-Norm-ReLU:3x3 (512) drop 0.2 | Conv-Norm-ReLU:3x3 (100)<br>Conv-Norm-ReLU:3x3 (100)<br>Conv-Norm-ReLU:3x3 (100)<br>Max-pool 2x2<br>Conv-Norm-ReLU:3x3 (200)<br>Conv-Norm-ReLU:3x3 (200)<br>Conv-Norm-ReLU:3x3 (200)<br>Max-pool 2x2<br>Conv-Norm-ReLU:3x3 (300) drop 0.2<br>Conv-Norm-ReLU:3x3 (300) drop 0.2<br>Conv-Norm-ReLU:3x3 (300) drop 0.2<br>Max-pool 1x2<br>Conv-Norm-ReLU:3x3 (400) drop 0.2<br>Conv-Norm-ReLU:3x3 (400) drop 0.2<br>Conv-Norm-ReLU:3x3 (400) drop 0.2<br>Max-pool 2x1<br>Conv-Norm-ReLU:3x3 (512) drop 0.2 |
| **Accuracy** | 32.48 | 32.48 | 35.91 |

TABLE III. PERFORMANCE OF END-TO-END SYSTEM TRAINED BY DIFFERENT TRAINING SETS.

| **Training set** \ **Performance** | **ExpRate (%)** | |
|---|---|---|
| | **CROHME 2014** | **CROHME 2016** |
| **CROHME training set** | 39.76 | 36.27 |
| **CROHME-Distortion** | 41.48 | 37.14 |
| **CROHME-Decomposition** | 47.67 | 42.63 |
| **CROHME-Hybrid** | 48.78 | 45.60 |

We investigate the performance of the end-to-end recognition system on "seen" and "unseen" sets by the following experiment, where the seen set stores HMEs whose ground-truths are in the training set, i.e., different HME patterns of the same MEs are used for training, while the unseen set stores HMEs whose ground-truths are not in the training set. We separated offline HMEs in the testing set to "seen" and "unseen" sets, and considered two conditions: exact match, when all the symbols and structures of two HME completely match, and structural match, when structures of two offline HMEs match while symbols do not match. Then, we calculate ExpRate of the best end-to-end recognition system (trained on CROHME Hybrid) on the "seen" and "unseen" sets. The number of HMEs and ExpRate for the "seen" and "unseen" sets in CROHME 2014 and 2016 are listed in Table IV. We found that the performance of the end-to-end recognition system on the "seen" set is better than on the "unseen" set. However, the recognition system is able to recognize unseen HMEs (Exact match: 47.59% and 44.81 on the CROHME 2014, 2016 test set, respectively; Structural match: 31.91% and 31.55% on the CROHME 2014, 2016 test set, respectively). This shows that the recognition system is able to learn elements of an HME such as symbols and relations between symbols, and is able to recognize new structures that it did not learn from the training dataset

TABLE IV. PERFORMANCE OF THE BEST END-TO-END RECOGNITION SYSTEM ON SEEN AND UNSEEN SETS OF CROHME 2014 AND 2016 TEST SET.

| **Testing set** \ **Performance** | | Exact match | | Structural match | |
|---|---|---|---|---|---|
| | | Seen set | Unseen set | Seen set | Unseen set |
| **CROHME 2014 test set** | **# of HMEs** | 32 | 954 | 538 | 448 |
| | **ExpRate (%)** | 84.38 | 47.59 | 62.83 | 31.91 |
| **CROHME 2016 test set** | **# of HMEs** | 20 | 1127 | 510 | 637 |
| | **ExpRate (%)** | 90.00 | 44.81 | 63.13 | 31.55 |

Figure 7 illustrates the recognition process of the end-to-end recognition systems trained on CROHME-Hybrid for a given offline HME. At every step, the end-to-end recognition system generates a symbol. The attention probabilities are visualized in red and the predicted symbols are shown on the right of HMEs. A dark red color mean a high probability of attention while a light red hue means a low probability. We can see that the attention on symbols is dark red while the attention on hidden symbols like "{" and "}" is lighter.

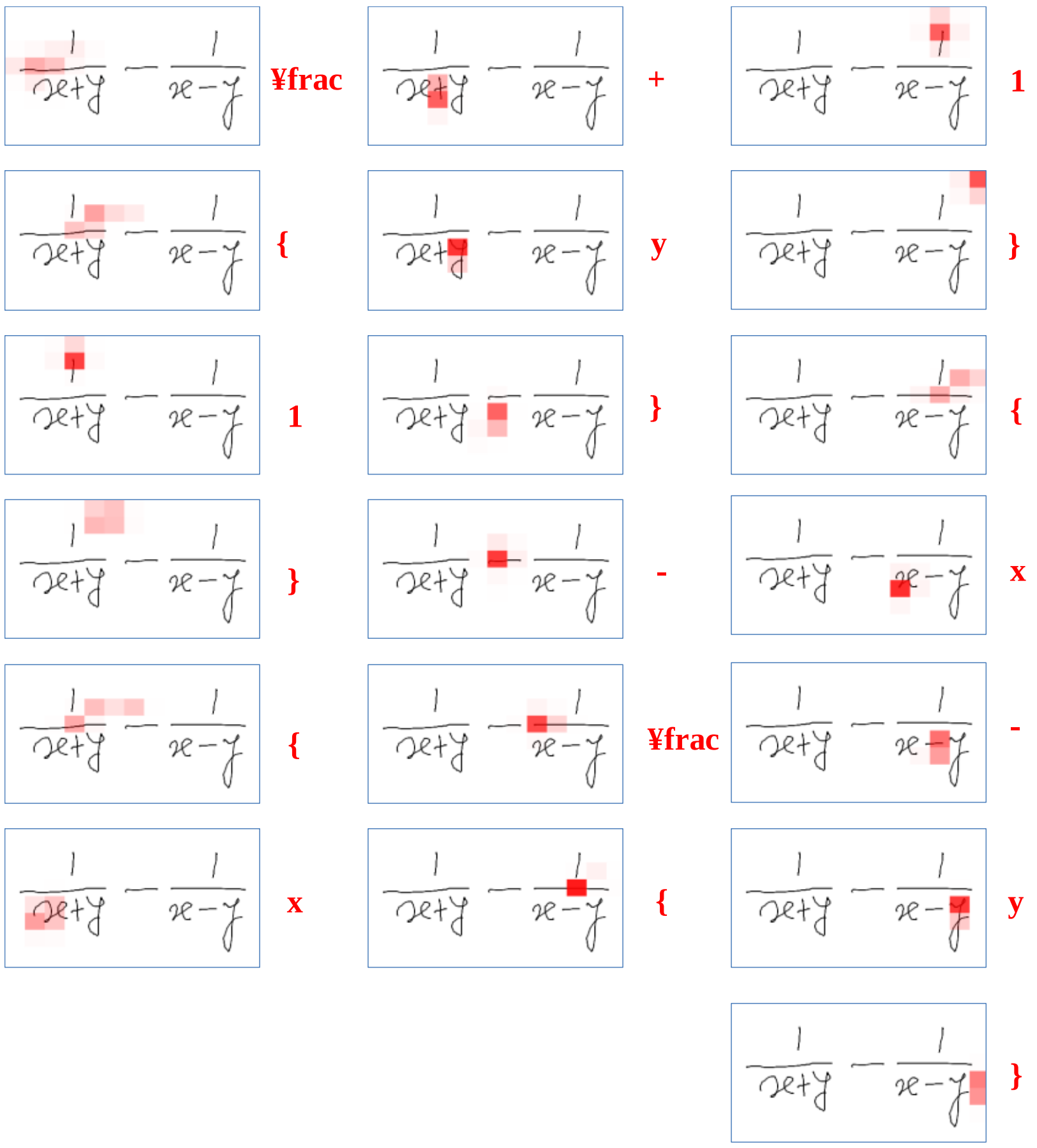

**Fig. 7.** Illustration of the recognition process

Figure 8 shows examples recognized correctly and incorrectly by the end-to-end recognition system trained on CROHME-Hybrid (recognition results are shown below the HMEs). Recognition errors frequently occur for ambiguous symbols (e.g. “B” is recognized as “$\beta$” ) and under-segmentation (e.g. two symbols “,” and “0” are recognized as “3”) as shown in Figure 8.b.

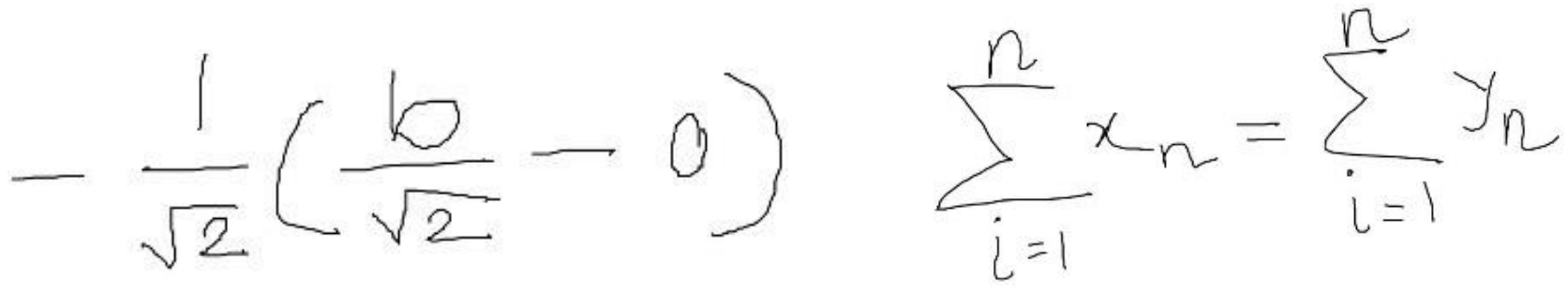

$$\frac{-1}{\sqrt{2}}\left(\frac{b}{\sqrt{2}}-0\right) \qquad \sum_{i=1}^{n} x_n = \sum_{i=1}^{n} y_n$$

a). Correct recognition

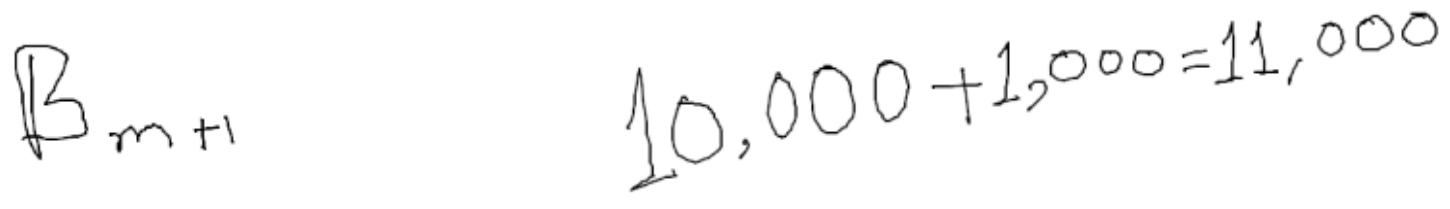

$$\boldsymbol{\beta_{m+1}} \qquad 1\,0\,,\,0\,0\,0\,+\,1\,3\,0\,0\,=\,1\,1\,.\,0\,0\,0$$

b) Recognition errors

**Fig. 8.** Examples recognized correctly and incorrectly by the best end-to-end recognition system.

## 4.4 Comparison with the state-of-the-art

We compared the best recognition system trained on CROHME Hybrid and others on CROHME 2014 and 2016. The results on CROHME 2014 are shown in Table V. Systems I to VII, which are based on Context Free Grammar, were submitted to the CROHME 2014 competition. They are described in [6]. MyScript’s system (system III), which uses a huge number of in-house training patterns, was ranked top for ExpRate. Excepting MyScript’s system, Avaro’s system (system I) achieved the best result using only the CROHME training dataset. For ExpRate, our best recognition system (listed at the bottom) is better than Alvaro’s system, but worse than Myscript's system. The columns denoted by < 1, <2, <3 show the recognition accuracies, when 1 to 3 errors, respectively, in symbols or relations are permitted. For these measurements, compared to the Myscript’s system, our best recognition system is a little worse for <1 and <2, and better for < 3. WYGIWYS, WAP, MSD and our recognition system are based on the encoder-decoder model with attention. WYGIWYS and our recognition system employ a CNN feature extraction, a BLSTM row encoder, and a unidirectional LSTM decoder with a classic attention model. WAP employs an encoder based on the VGG architecture and its decoder is a unidirectional GRU equipped with a coverage-based attention model. MSD is the improved version of WAP, which employs Multi-Scale Attention with a DenseNet encoder. WAP and MSD are combined from 5 different models. For ExpRate, our best recognition system is better than WYSIWYG, WAP and worse than MSD. For <1, <2, <3 error measurements, our best recognition system outperforms all of them.

TABLE V. COMPARISION OF END-TO-END MODEL AND RECOGNITION SYSTEMS ON CROHME 2014 (%)

| Measure / Method | LM | Ensemble | Exp Rec | <1 | <2 | <3 |
|---|---|---|---|---|---|---|
| I | NO | NO | 37.22 | 44.22 | 47.26 | 50.20 |
| II | NO | NO | 15.01 | 22.31 | 26.57 | 27.7 |
| III | YES | NO | **62.68** | **72.31** | **75.15** | 76.88 |
| IV | NO | NO | 18.97 | 28.19 | 32.35 | 33.37 |

| Measure / Method | LM | Ensemble | Exp Rec | <1 | <2 | <3 |
|---|---|---|---|---|---|---|
| I | NO | NO | 37.22 | 44.22 | 47.26 | 50.20 |
| II | NO | NO | 15.01 | 22.31 | 26.57 | 27.7 |
| III | YES | NO | **62.68** | **72.31** | **75.15** | 76.88 |
| IV | NO | NO | 18.97 | 28.19 | 32.35 | 33.37 |
| V | NO | NO | 18.97 | 26.37 | 30.83 | 32.96 |
| VI | NO | NO | 25.66 | 33.16 | 35.90 | 37.32 |
| VII | NO | NO | 26.06 | 33.87 | 38.54 | 39.96 |
| WYGIWYS | NO | NO | 39.96 | N/A | N/A | N/A |
| WAP | NO | YES | 44.4 | 58.4 | 62.2 | 63.1 |
| MSD | NO | YES | 52.8 | 68.1 | 72.0 | 72.7 |
| Our best recognition system | NO | NO | 48.78 | 66.13 | 73.94 | **79.01** |

The results on CROHME 2016 are listed in Table VI. The first five systems participated in CROHME 2016. MyScript's system, using in-house training patterns, was still ranked top for ExpRate. WIRIS system is the improved version of Alvaro's system which integrated a language model. It was given the best result award using only the CROHME training dataset. For ExpRate, our system is inferior to Myscript, but competitive with WIRIS and MSD. For the <1, <2, <3 errors measurements, our system is better than WIRIS and worse than only Myscript (<1, <2) and MSD (<1).

TABLE VI. COMPARISION OF END-TO-END MODEL AND RECOGNITION SYSTEMS ON CROHME 2016 (%)

| Measure / Method | LM | Ensemble | Exp Rec | <1 | <2 | <3 |
|---|---|---|---|---|---|---|
| MyScript | NO | NO | **67.65** | 75.59 | 79.86 | - |
| WIRIS | NO | NO | 49.6 | 60.4 | 64.7 | - |
| TOKYO | NO | NO | 43.9 | 50.9 | 53.7 | - |
| Sao Paolo | NO | NO | 33.4 | 43.5 | 49.2 | - |
| Nantes | NO | NO | 13.3 | 21.0 | 28.3 | - |
| WAP | NO | YES | 42.0 | 55.1 | 59.3 | 60.2 |
| MSD | NO | YES | 50.1 | 63.8 | 67.4 | 68.5 |
| Our best recognition system | NO | NO | 45.60 | 62.25 | 70.44 | **75.76** |

The important observation is that gaps between the correct and the <1, <2, <3 error measurements of our system are higher than those of others on both CROHME 2014 and 2016. This shows that our system has room for further improvements in the language model, ensembles, and the attention model.

- Language model: we can employ the N-gram model to revise incorrect recognition.
- Ensemble: we can combine multiple end-to-end recognition systems to improve the recognition rate.
- Attention model: we can employ the coverage model, temporal attention presented in [9 - 10] to improve the attention model.

## 5 CONCLUSION

In this paper, we have proposed three pattern generation strategies for recognizing HMEs: distortion, decomposition, and hybrid. We also presented an end-to-end HME-recognition system based on the encoder-decoder and attention model. The efficiency of the proposed strategies were demonstrated through experiments. The recognition rate is improved when we employ the pattern generation strategies. The decomposition strategy is more effective than the distortion strategy because it generates more structural variations. The best accuracy is achieved with the hybrid strategy: 48.78% and 45.60% on CROHME 2014 and 2016, respectively. Our best recognition system is competitive with the state-of-the-art recognition systems, but our system is simpler and has a scope for much further improvement. Our generated data is openly available for the research community.

## ACKNOWLEDGMENT

This work was supported by the JSPS fellowship under the contract number 15J08654.